\documentclass[preprint]{IEEEtran}

\usepackage{url}

\usepackage{fancyhdr}
\usepackage{float}
\usepackage{multirow}
\usepackage{graphicx}
\usepackage{xcolor}
\usepackage{array}
\usepackage{subcaption}
\usepackage{hyperref}
\usepackage[linesnumbered, ruled]{algorithm2e}
\usepackage{amsmath} % some math-related packages, not sure which of them are necessary
\usepackage{amssymb}
\usepackage{amsfonts}
\usepackage{orcidlink}
\begin{document}

\title{Enhanced 3D Object Detection via Diverse Feature Representations of 4D Radar Tensor}

\author{Seung-Hyun Song\orcidlink{0009-0001-4813-779X},~\IEEEmembership{Student~Member,~IEEE}, Dong-Hee Paek\orcidlink{0000-0003-0008-3726},~\IEEEmembership{Student~Member,~IEEE}, Minh-Quan Dao\orcidlink{0009-0001-4132-2159},~\IEEEmembership{Member,~IEEE}, Ezio Malis\orcidlink{0000-0002-6584-6790},~\IEEEmembership{Member,~IEEE}, and Seung-Hyun~Kong\orcidlink{0000-0002-4753-1998},~\IEEEmembership{Senior~Member,~IEEE}% 

\thanks{Manuscript received [Submission Date]; revised [Revision Date]; accepted [Acceptance Date]. Date of publication [Publication Date]; date of current version 14 December 2024. This work was supported by the National Research Foundation of Korea(NRF) grant funded by the Korea government(MSIT) (No. 2021R1A2C3008370). \textit{(Corresponding authors: Seung-Hyun Kong.)}}
\thanks{Seung-Hyun Song is with the Graduate School of Advanced Security Science and Technology, Korea Advanced Institute of Science and Technology, Daejeon, Korea, 34051 (e-mail: shyun@kaist.ac.kr)}
\thanks{Dong-Hee Paek and Seung-Hyun Kong are with the CCS Graduate School of Mobility, Korea Advanced Institute of Science and Technology, Daejeon, Korea, 34051 (e-mail: donghee.paek@kaist.ac.kr; skong@kaist.ac.kr)}
\thanks{Minh-Quan Dao and Ezio Malis are with Centre Inria d'Univeristé Côte d'Azur (e-mail: minh-quan.dao@inria.fr; ezio.malis@inria.fr)} 
}

% The paper headers
% \markboth{IEEE ROBOTICS AND AUTOMATION LETTERS, Vol. , No. , December 2024}{Song \MakeLowercase{\textit{et al.}}: A Novel Multi-Teacher Knowledge Distillation for Real-Time Object Detection using 4D Radar}

\maketitle

\begin{abstract}
Recent advances in automotive four-dimensional (4D) Radar have enabled access to raw 4D Radar Tensor (4DRT), offering richer spatial and Doppler information than conventional point clouds. While most existing methods rely on heavily pre-processed, sparse Radar data, recent attempts to leverage raw 4DRT face high computational costs and limited scalability.
To address these limitations, we propose a novel three-dimensional (3D) object detection framework that maximizes the utility of 4DRT while preserving efficiency. Our method introduces a multi-teacher knowledge distillation (KD), where multiple teacher models are trained on point clouds derived from diverse 4DRT pre-processing techniques, each capturing complementary signal characteristics. These teacher representations are fused via a dedicated aggregation module and distilled into a lightweight student model that operates solely on a sparse Radar input.
Experimental results on the K-Radar dataset demonstrate that our framework achieves improvements of 7.3\% in AP$_{3D}$ and 9.5\% in AP$_{BEV}$ over the baseline RTNH model when using extremely sparse inputs. Furthermore, it attains comparable performance to denser-input baselines while significantly reducing the input data size by about 90 times, confirming the scalability and efficiency of our approach.

\end{abstract}


\IEEEpeerreviewmaketitle

\section{Introduction} \label{sec:intro}

\IEEEPARstart{A}{utonomous} navigation relies on accurate object detection in three-dimensional (3D) space to precisely localize nearby road users and ensure safe operation. Robust performance under all weather conditions is essential for ensuring overall safety \cite{geiger2012we,caesar2020nuscenes,mirza2021robustness}.
While LiDAR has traditionally been the primary sensing modality for this task due to its ability to accurately measure 3D distances, it struggles to function reliably in adverse conditions such as snow, rain, and fog, owing to its reliance on light signals that are harmless to human eyes \cite{bijelic2018benchmark}.
In contrast, Radar sensors remain robust under adverse weather conditions due to their use of electromagnetic waves. 
Recent advancements in Radar technology have led to the development of four-dimensional (4D) Radars, which augment traditional Radar measurements—range, azimuth, and Doppler velocity—with elevation information.
This enhancement lifts Radar perception into the full 3D space, positioning 4D Radar as a promising sensing modality for robust 3D object detection \cite{palffy2022multi, zheng2022tj4dradset, kradar}.

\begin{figure}[t!]
    \centerline{\includegraphics[width=0.95\linewidth]{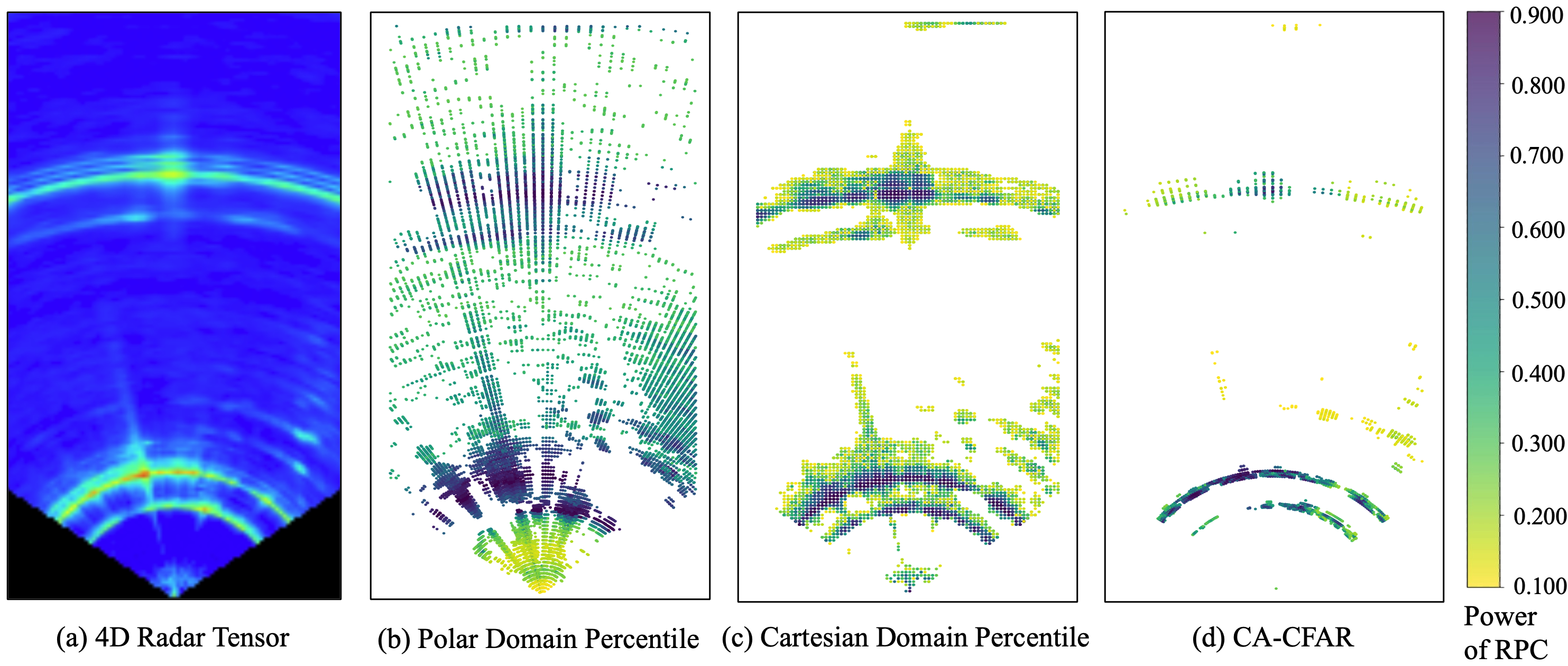}}
    \caption{
        Comparison of 4D Radar point clouds based on different pre-processing techniques: (a) original output from 4DRT, (b) Fixed percentile filtering in the polar domain, (c) Fixed percentile filtering in the cartesian domain with interpolation, and (d) CA-CFAR filtering.
        }
    \label{fig.various rpc}
\end{figure}

%%문제점
The raw measurements of 4D Radar can take various forms—such as microwave signals, digital signals, range-Doppler maps, 4DRT, and point clouds—depending on the signal processing stage. Due to inherent Radar characteristics, such as low signal-to-noise ratio (SNR), multipath reflections, and phase noise, Radar signals undergo complex processing steps involving distinct assumptions and filtering techniques. As a consequence, the same environmental scene can yield significantly different data distributions. For instance, as illustrated in Fig.\ref{fig.various rpc}, applying different pre-processing techniques (e.g., polar vs. Cartesian coordinate conversions or varying filtering logics and density thresholds) to identical 4DRT inputs results in markedly diverse point distributions and semantic characteristics. This inherent variability and inconsistency pose significant challenges in designing perception systems that robustly learn consistent and informative feature representations from 4D Radar data.

%% 기존 접근법
To overcome this, two main research directions have emerged:
(1) designing architectures that more effectively extract semantic and contextual information from sparse Radar inputs \cite{rpfa, pointpillars, smurf, mvfan, smiformer, radarmfnet};
(2) exploiting richer raw measurements such as 4DRT, aiming to capture latent features in the early signal domain \cite{fftradnet, adcnet, t-fftradnet, fent2024dpft, rtnh+}.

However, existing approaches remain insufficient in addressing this fundamental challenge. Methods based on sparse point clouds typically rely on a single pre-processed representation that is heavily filtered and lossy, making them incapable of capturing the diverse and complementary signal characteristics inherent in 4DRT. This narrow representation scope limits the model’s robustness under varying pre-processing assumptions and noise conditions. Conversely, approaches that directly process raw 4DRT aim to preserve richer signal information, but their computational and memory demands are prohibitively high for real-time or resource-constrained deployment. As such, neither direction effectively mitigates the inherent variability and inconsistency introduced during Radar signal processing—leaving a critical gap in robust and scalable Radar-based perception.

To overcome these limitations, we propose a 3D object detection framework that leverages the diverse and complementary information embedded in 4DRT without incurring the high computational costs of direct raw processing. Our method is built on a multi-teacher KD paradigm, where multiple teacher models are trained on point clouds derived from different 4DRT pre-processing techniques—each capturing distinct semantic and structural characteristics. These teacher representations are fused via a dedicated aggregation module and distilled into a lightweight student model that operates solely on sparse Radar inputs. This enables the student to effectively learn rich 4DRT features despite limited input density, maximizing both the informativeness and efficiency of Radar-based perception.

To summarize, our key contributions are:
\begin{itemize}
    \item We present the first multi-teacher KD that leverages diverse feature representations of the 4DRT, obtained through multiple pre-processing techniques, to enhance 3D object detection from sparse Radar inputs.
    
    \item By leveraging the proposed aggregation module—designed to fuse various 4DRT representations—the student model effectively distills rich features and achieves competitive 3D object detection using only sparse inputs with minimal computational cost.

    \item Comprehensive experiments on the large-scale K-Radar dataset \cite{kradar}, results show a 7.3\% improvement in AP$_{3D}$ over the Baseline RTNH model trained on the same sparse 99.9th percentile input.
    
\end{itemize}

%%%%%%%%%%%%%%%%%%%%%%%%%%%%%%%%%%%%%%%%%%%%%%%%%%%%%%%%%%%%%%%%%%%%%%%%%%%%%%%%%%%%%%%%%%%%%%%%%%%%%%%%%%%%%%%%%%%%%
%%%%%%%%%%%%%%%%%%%%%%%%%%%%%%%%%%%%%%%%%%%%%%%%%%%%%%%%%%%%%%%%%%%%%%%%%%%%%%%%%%%%%%%%%%%%%%%%%%%%%%%%%%%%%%%%%%%%%
%%%%%%%%%%%%%%%%%%%%%%%%%%%%%%%%%%%%%%%%%%%%%%%%%%%%%%%%%%%%%%%%%%%%%%%%%%%%%%%%%%%%%%%%%%%%%%%%%%%%%%%%%%%%%%%%%%%%%

%%%%%%%%%%%%%%%%%%%%%%%%%%%%%%%%%%%%%%%%%%%%%%%%%%%%%%%%%%%%%%%%%%%%%%%%%%%%%%%%%%%%%%%%%%%%%%%%%%%%%%%%%%%%%%%%
%%%%%%%%%%%%%%%%%%%%%%%%%%%%%%%%%%%%%%%%%%%%%%%%%%%%%%%%%%%%%%%%%%%%%%%%%%%%%%%%%%%%%%%%%%%%%%%%%%%%%%%%%%%%%%%%
\section{Related Works}
\subsection{4D Radar pre-processing Techniques}
Since the representational quality of 4D Radar point clouds is highly affected by pre-processing, various techniques have been actively investigated.
Traditional 4D Radar pre-processing techniques are generally categorized into fixed thresholding and Constant False Alarm Rate (CFAR) techniques. Fixed thresholding selects points based on a global or percentile-based power threshold, while CFAR adaptively determines local thresholds using the statistical properties of surrounding cells.

RTNH\cite{kradar} introduced a method that first transforms the 4DRT into Cartesian coordinates via interpolation, followed by percentile-based filtering. This approach preserves spatial consistency and enables the capture of meaningful low-power signals.
Building on this, RTNH+\cite{rtnh+} proposed a CFAR-based two-level pre-processing (TLP) technique. It first applies a coarse CA-CFAR with a high threshold to retain as many relevant measurements as possible. Then, a second stage applies a range-wise percentile filter with a lower threshold along the azimuth axis to remove spurious points and enhance data reliability.

In this work, we aim to extract diverse feature representations from 4DRT using multiple pre-processing techniques and leverage them to improve 3D object detection performance.

\begin{figure*}[h]
    \centerline{\includegraphics[width=0.9\linewidth]{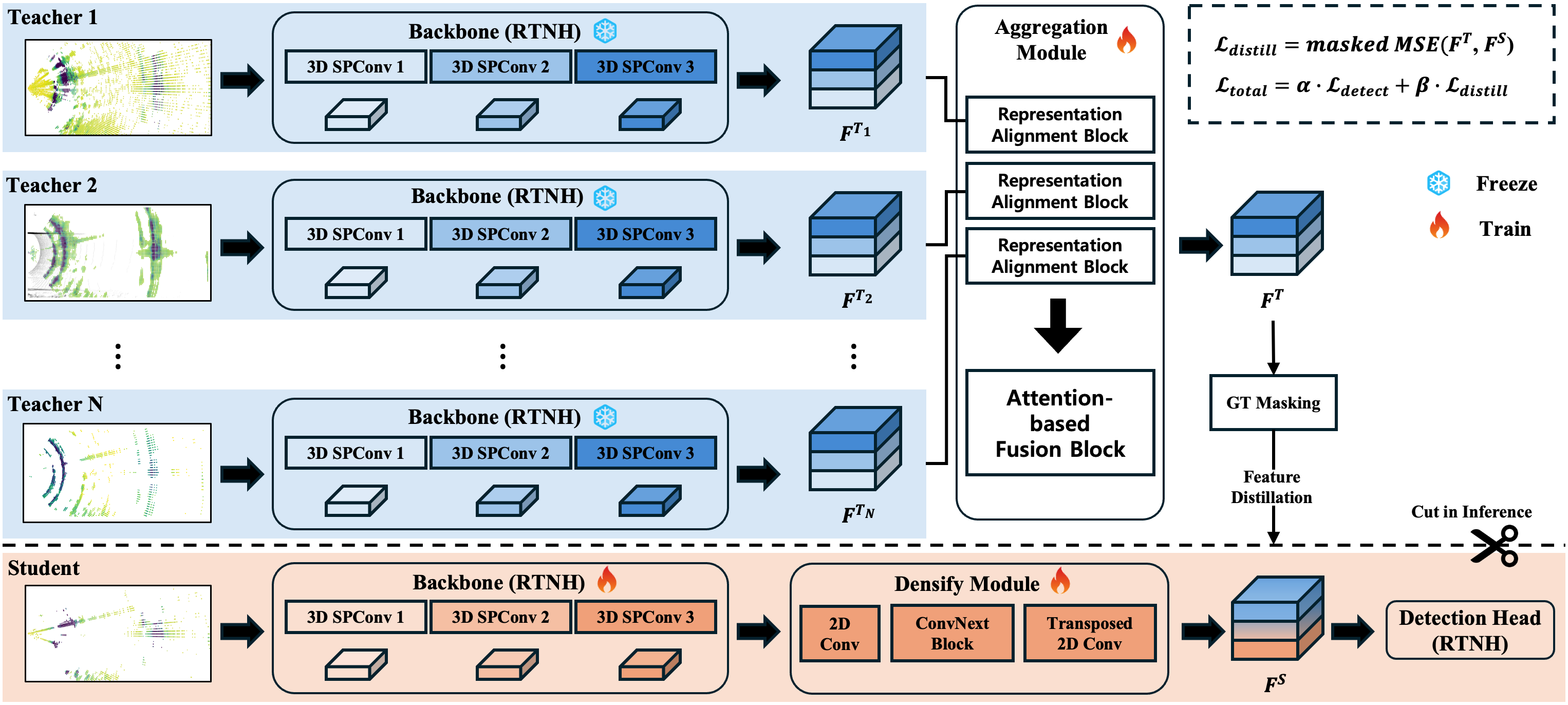}}
    \caption{
        \textbf{4DR-MR}: Overall architecture of the proposed 3D object detection framework utilizing multi-representations of 4D Radar. Features from multiple teacher networks, each using different Radar pre-processing, are fused and distilled into a compact student model for efficient inference.  
    }
    \label{fig.architecture}
\end{figure*}

\begin{figure}[ht]
    \centerline{\includegraphics[width=0.99\linewidth]{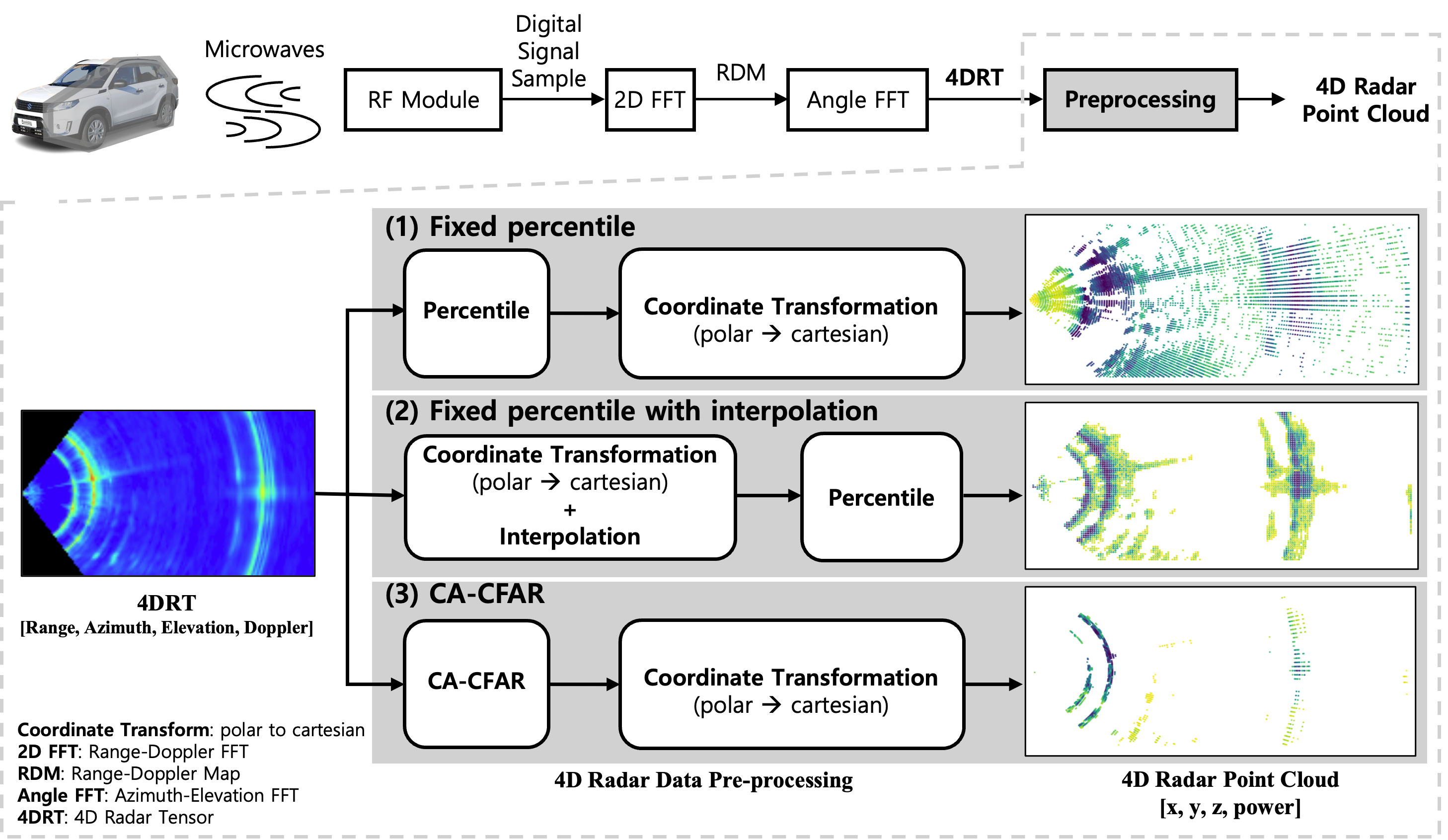}}
    \caption{
        Radar processing pipeline: Microwave signals are digitized, then transformed via two FFTs into a dense 4D tensor. A pre-processing step then converts it to a sparse point cloud.
    }
    \label{fig.Radar processing}
\end{figure}

\subsection{3D Object Detection Architectures with 4D Radar}
Recent advances in 3D object detection using 4D Radar data have primarily focused on designing architectures that effectively capture semantic and contextual information from sparse point clouds.

RPFA-Net \cite{rpfa} builds upon the Pillar Feature Net from PointPillars \cite{pointpillars}, incorporating a self-attention module \cite{attention} to enhance the modeling of relationships among Radar points within each pillar. Similarly, SMURF \cite{smurf} extends the PointPillar architecture by enriching pillar-based features with density-based features extracted via kernel density estimation, thereby mitigating the effects of point cloud sparsity.
RTNH \cite{kradar} utilizes 4DRT to demonstrate that the density of Radar point clouds significantly affects detection performance \cite{enhancedkradar}, and employs a sparse 3D convolutional architecture to effectively integrate height-related features.
MVFAN \cite{mvfan} adopts a multi-branch architecture that extracts features from both bird's-eye view (BEV) and cylindrical view representations, while SMIFormer \cite{smiformer} further generalizes this idea by leveraging BEV, front view (FV), and side view (SV) representations, which are fused using multi-view interaction transformers.
To account for dynamic object motion, RadarMFNet \cite{tan20223} aggregates Radar point clouds across time, leveraging Doppler velocity to compensate for object displacement and obtain denser input representations.
Radar PillarNet \cite{rcfusion} generates Radar pseudo images using a pillar-based encoding scheme, where spatial, velocity, and intensity features are separately processed to improve BEV feature learning.

In parallel, several studies have shifted toward directly processing early-stage Radar signals.
The early work FFT-RadNet \cite{fftradnet} performs object detection directly on the range-Doppler map (RDM). As RDMs lack explicit azimuth information, FFT-RadNet implicitly estimates azimuth within its latent feature space by learning an axis that encodes azimuthal variation.
ADCNet \cite{adcnet} and T-FFTRadNet \cite{t-fftradnet} further extend this direction by processing raw ADC signals. Leveraging learnable versions of the discrete Fourier transform, they introduce neural signal processing modules that transform the ADC signal into a range-azimuth-Doppler latent representation.
More recently, DPFT \cite{fent2024dpft} takes the 4DRT as input and projects it onto range–azimuth and azimuth–elevation planes for feature extraction. The resulting feature maps are then fused with camera data using deformable attention \cite{zhu2021deformable}.

Unlike prior works that rely on a single representation, our framework distills knowledge from multiple representations—each capturing complementary aspects of the 4DRT. Our approach aims to bridge the gap between sparse point cloud-based detection and raw signal-level exploitation, achieving high semantic understanding while maintaining computational efficiency.

%%%%%%%%%%%%%%%%%%%%%%%%%%%%%%%%%%%%%%%%%%%%%%%%%%%%%%%%%%%%%%%%%%%%%%%%%%%%%%%%%%%%%%%%%%%%%%%%%%%%%%%%%%%%%%%%
%%%%%%%%%%%%%%%%%%%%%%%%%%%%%%%%%%%%%%%%%%%%%%%%%%%%%%%%%%%%%%%%%%%%%%%%%%%%%%%%%%%%%%%%%%%%%%%%%%%%%%%%%%%%%%%%

\begin{figure*}[t]
    \centerline{\includegraphics[width=0.9\linewidth]{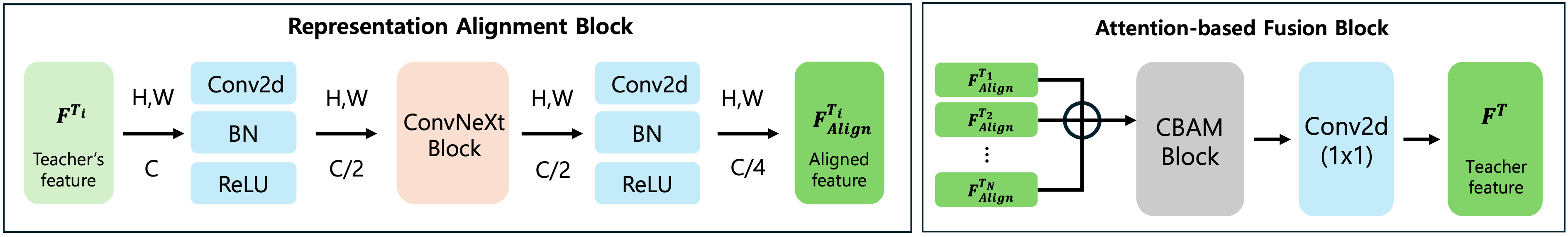}}
    \caption{
         Aggregation module: The module consists of a Representation Alignment Block that refines each teacher feature and an Attention-based Fusion Block that adaptively integrates the aligned features.  
    }
    \label{fig.aggregation}
\end{figure*}

\section{Proposed Approach} \label{sec:methodology}
We propose a 3D object detection framework called 4D Radar Multi-Representation (4DR-MR). 
This section first presents an overview of the main components and data flow of our framework, 
followed by a detailed description of each module.

\subsection{Overview of 4DR-MR} \label{sec:overall Framework}
The 4DR-MR framework adopts a multi-teacher KD, 
in which multiple teacher models and a single student model are jointly trained 
to transfer knowledge across diverse Radar representations (see Fig. \ref{fig.architecture}).

Each teacher model receives its own Radar point cloud input and extracts BEV features $F^{T_i}$ using a non-shared RTNH backbone \cite{kradar}. 
The resulting features from all teachers are passed into the Aggregation Module, which fuses them into a unified representation $F^T$.
This aggregated teacher feature $F^T$ supervises the student model, which shares the same backbone architecture but operates on a separate sparse Radar input. 
The student feature $F^S$, refined by the Densify Module, is then compared to the fused teacher features for distillation.

The entire framework is trained with object detection loss $\mathcal{L}{\text{detect}}$ and feature-level distillation loss $\mathcal{L}{\text{distill}}$, where $\mathcal{L}_{\text{distill}}$ is calculated using ground truth-masked features to focus learning on object regions.
At inference time, only the student model is used, enabling efficient 3D object detection without the overhead of teacher models or the aggregation module.

\subsection{Radar Pre-processing for 4DR-MR}
We implement and analyze three pre-processing techniques to exploit the rich information embedded in the 4DRT.
In addition to the type of processing applied, we also consider the impact of point cloud density by adjusting percentile thresholds or filtering criteria within each strategy.
\subsubsection{Fixed Percentile in the Polar Domain}
This method extracts Radar point clouds by directly applying percentile-based thresholding to the power tensor computed from the 4DRT. Specifically, the Doppler axis of the 4DRT is first collapsed by averaging to obtain a 3D power volume. A global power threshold is then set using the $r$-th percentile value across all power values. Each discrete [azimuth, range, elevation] coordinate is transformed to Cartesian coordinates, and points exceeding the threshold are retained. As illustrated in Algorithm \ref{alg:preproc}, this approach preserves the Radar's native polar geometry and returns sparse but strong detections with minimal computational cost. However, due to the lack of spatial regularity, the resulting point cloud is less compatible with voxel-based convolutional architectures.

\begin{algorithm}
    \DontPrintSemicolon
    \SetNoFillComment
    \SetKwInOut{Input}{Input}
    \SetKwInOut{Output}{Output}
    \Input{
        $\mathbf{4DRT}$ $\in \mathbb{R}^4$ \small(azimuth, range, elevation, Doppler), 
        \\
        percentile $r$ $\in \mathbb{R}$
    }
    \Output{Radar point cloud $\mathbb{P} = \left\{ \mathbf{p}_i = \left[x, y, z, power\right] \right\}$}

    \BlankLine

    \tcc{\scriptsize Power Mapping}
    $\mathbf{power}$ = \textrm{mean}$\left(\mathbf{4DRT}, ~\textrm{dim=Doppler}\right)$  
    
    \BlankLine

    $power\_threshold$ = $r$ percentile of $\mathbf{power}$ 
    
    \BlankLine
    \tcc{\scriptsize Thresholding and Coordinate Transformation}
    
    $\mathbb{P} = \emptyset$

    \For{\footnotesize each discrete coordinate [azimuth, range, elevation] of $\mathbf{power}$}{
        $pw = \mathbf{power}$[azimuth, range, elevation]

        $azi$, $rg$, $ele$ = \text{discrete\_to\_continuous}([azimuth, range, elevation])

        \If{$pw \geq power\_threshold$}{
            $x = rg \cdot \cos(ele) \cdot \cos(azi)$

            $y = rg \cdot \cos(ele) \cdot \sin(azi)$

            $z = eg \cdot \sin(ele)$

            \text{add} $\mathbf{p} = [x, y, z, power]$ to $\mathbb{P}$
        }
    }
    \caption{Fixed percentile-based pre-processing}
    \label{alg:preproc}
\end{algorithm}

\subsubsection{Fixed Percentile in the Cartesian Domain with Interpolation}

This method follows the pre-processing technique proposed in RTNH \cite{kradar}.
The 4DRT is first interpolated into a regular Cartesian voxel grid using bilinear interpolation, transforming the data from its native polar coordinate system.
Subsequently, percentile-based filtering is applied in the voxel space, using the same settings as RTNH \cite{kradar}.
This approach aligns well with voxel-based 3D CNNs and BEV detectors by ensuring spatial regularity.
Although the interpolation step incurs additional computational overhead, it enables structural continuity and uniform resolution, making it particularly suitable for dense perception pipelines.

\subsubsection{CA-CFAR in the Polar Domain}

This method applies the CA-CFAR algorithm to the 4DRT in its native polar coordinate system. It retains points whose power values exceed a locally computed threshold based on surrounding cells, effectively suppressing background noise and emphasizing strong reflections, particularly along object boundaries.
While this approach improves precision by reducing clutter, it often removes low-power returns within object interiors, leading to sparser and less complete point clouds that are suboptimal for learning-based 3D detection models.

\subsection{Aggregation Module}  \label{sec:Aggregationmodule}
To effectively transfer the various representation knowledge of 4DRT from multiple teacher models to a student model, we design a novel feature aggregation mechanism composed of two key components: (1) a Representation Alignment Block for teacher-specific feature transformation, and (2) an Attention-based Fusion Block for adaptive integration of multi-teacher features.

\subsubsection{Representation Alignment Block}
Each teacher model provides distinct feature representations that may differ in terms of scale, semantic content, and distribution. To address this heterogeneity, we introduce a Representation Alignment Block that independently processes the feature of each teacher $F^{T_i}$.
As illustrated in Fig.\ref{fig.aggregation} (left), the Representation Alignment Block consists of three stages: an initial 1x1 convolution with batch normalization and ReLU activation to compress the feature channels, a ConvNeXt inspired block to enhance channel-wise interactions and nonlinear representation, and a final projection to align the output to a common dimensionality C/4. This sequential transformation harmonizes the characteristics of the teacher features, enabling a more effective distillation of knowledge into the student model. Formally, for a teacher feature $F^T \in \mathbb{R}^{C \times H \times W}$, the aligned feature $F^{T_i}_{Align} \in \mathbb{R}^{C/4 \times H \times W}$ is obtained by:
\[
F^{T_i}_{Align} = \text{KnowledgeAlignmentBlock}(F^{T_i})
\]
This alignment ensures that teacher features, though diverse in origin, are semantically harmonized and ready for fusion.

\subsubsection{Attention-based Fusion Block}
The aligned teacher features $F^{T_i}_{Align}$ are concatenated along the channel dimension and passed through the Attention-based Fusion Block. As shown in Fig.\ref{fig.aggregation} (right), this block first applies a CBAM (Convolutional Block Attention Module) to the concatenated features, selectively enhancing informative channels and spatial regions. Subsequently, a 1x1 convolution is employed to integrate the features into a final aggregated representation. 
The overall fusion operation can be formulated as:
\[
F^{T} = \text{Conv}_{1\times1}(\text{CBAM}(\text{Concat}(F^{T_i}_{Align}))),  i\in \{1,2,...,N\}
\]
This fusion strategy adaptively highlights the most informative aspects of multi-teacher knowledge, improving the efficiency and effectiveness of student learning.

\subsection{Densify Module} \label{sec:densifymodule}
In our framework, features extracted from the teacher and student backbones exhibit a density mismatch due to differences in their input point clouds.
While the teacher models, benefiting from occupancy leaking and denser inputs, generate relatively dense features, the student model—operating on extremely sparse inputs—produces under-represented BEV features even after several convolutional stages.
This discrepancy poses a challenge for feature-level KD, as it leads to mismatched feature semantics and spatial structures.

To mitigate this issue, we extend the Densify Module originally proposed in \cite{sparse2dense}, incorporating a dual-pass encoder–decoder refinement scheme for enhanced feature reconstruction.
In our implementation, the student BEV feature first passes through an encoder–decoder block augmented with ConvNeXt-style residual layers.
The output is then refined again via a second pass of the same encoder–decoder structure, effectively deepening the feature densification process.
Finally, we fuse the densified features with the original sparse BEV feature using a residual connection, where separate fusion layers are applied to each.

The resulting feature is further processed by a sequence of convolutional layers to produce the final student BEV representation $F^S$, which is used in the distillation loss against the aggregated teacher feature.

\subsection{Loss Function} \label{sec:loss_function}
The loss function $L_{total}$ used to train the student model is a weighted sum of object detection loss $L_{detect}$ and KD loss $L_{distill}$

\begin{equation}
    L_{total} = \alpha\cdot L_{detect} + \beta\cdot L_{distill}
\end{equation}

Here, $\alpha$ and $\beta$ are hyperparameters that adjust the balance between object detection loss and KD loss.
They are set to 1 in our experiments.

$L_{distill}$ is the Mean Squared Error (MSE) between the aggregation of the teachers' feature $\mathbf{F}^T$ and the student's densified feature $\mathbf{F}^S$.
Since features in regions where objects exist are more important for detecting objects than background, we mask both $\mathbf{F}^T$ and $\mathbf{F}^S$ using ground truth when calculating the distillation loss.
Specifically, we transform each ground truth object into a 2D Gaussian and splat it to the BEV to obtain a heatmap. 
This heatmap is multiplied element-wise with $\mathbf{F}^T$ and $\mathbf{F}^S$ to realize the masking operation.

\begin{equation}
    L_{distill} = \text{MSE}(\text{mask}(\mathbf{F}^T), \text{mask}(\mathbf{F}^S))
\end{equation}

We use the loss function defined by RTNH \cite{kradar} as the detection loss $L_{detect}$.

\begin{table*}[h!]
\centering
\caption{Comparison with RTNH baselines using various pre-
processing techniques. \textbf{Bold} indicates the proposed method.}
\label{tab:RTNH_baseline}
\renewcommand{\arraystretch}{1.2} % 행 간격 조정
\setlength{\tabcolsep}{8pt} % 열 간격 조정
\begin{tabular}{cc|c|c|cccc}
\hline \hline
\multicolumn{2}{c|}{\multirow{2}{*}{Pre-processing}}                               & \multirow{2}{*}{\begin{tabular}[c]{@{}c@{}}Data {[}MB{]}\\ /Frame\end{tabular}} & \multirow{2}{*}{Network} & \multicolumn{2}{c}{Sedan}       & \multicolumn{2}{c}{Bus or Truck} \\ \cline{5-8} 
\multicolumn{2}{c|}{}                                                              &                                        &                          & $AP_{3D}$         & $AP_{BEV}$       & $AP_{3D}$          & $AP_{BEV}$         \\ \hline \hline
\multicolumn{1}{c|}{\multirow{4}{*}{(1) Fixed Percentile}} & \multirow{2}{*}{99.9} & \multirow{2}{*}{0.1}                   & \textbf{4DR-MR}          & \textbf{44.16} & \textbf{52.59} & \textbf{25.62}  & \textbf{29.02}  \\ \cline{4-8} 
\multicolumn{1}{c|}{}                                      &                       &                                        & RTNH$_{99.9}$               & 36.84          & 43.10          & 19.55           & 25.60           \\ \cline{2-8} 
\multicolumn{1}{c|}{}                                      & 90                    & 9                                      & RTNH$_{90}$                 & 43.86          & 53.36          & 27.62           & 36.80           \\ \cline{2-8} 
\multicolumn{1}{c|}{}                                      & 80                    & 18                                     & RTNH$_{80}$                 & 47.08          & 55.23          & 33.11           & 37.91           \\ \hline
\multicolumn{2}{c|}{\begin{tabular}[c]{@{}c@{}}(2) Fixed Percentile \\ with Interpolation\end{tabular}}                       & 5                                      & RTNH                     & 50.71          & 56.47          & 33.95           & 43.80           \\ \hline
\multicolumn{2}{c|}{(3) CA-CFAR}                                                   & 1                                      & RTNH$_{cfar}$               & 46.30          & 51.90          & 31.27           & 22.96           \\ \hline \hline
\end{tabular}
\end{table*}

\section{Experiments}
\label{sec:experiments}

\subsection{Experimental Setup}
\subsubsection{Implementation Details}
Due to the limitation of our hardware, we set the number of teachers $N$ to 3. To ensure that each teacher learns knowledge from a distinct representation, we adopt the following configurations for the Radar pre-processing techniques:
(1) For the fixed percentile method, we use two thresholds: 80th and 90th percentile to generate different levels of sparsity;
(2) For the fixed percentile with interpolation method, we follow the same setting as used in RTNH \cite{kradar};
(3) For the CA-CFAR method, we adopt the configuration used in the TLP algorithm from RTNH+ \cite{rtnh+}.

We use the same RTNH-based backbone for both teacher and student models. 
Implementation details of the Aggregation Module and the Densify Module, including layer configurations and attention operations, are provided in Appendix~\ref{appendix:architecture_detail}.

Both the teacher and student models are trained under the same settings as the baseline RTNH [7].
The training epoch, learning rate, and batch size are set to 30, 0.001, and 8, respectively.
All experiments are conducted on an NVIDIA RTX 3090 GPU.

\subsubsection{Datasets and Evaluation Metrics} 
We conduct our experiments on the K-Radar dataset \cite{kradar}, a multi-modal large-scale dataset collected under diverse driving environments, including adverse weather conditions. It provides access to dense 4DRT, which is crucial for our approach. The dataset contains 35,000 frames, evenly split into training and testing sets, and includes 100,000 annotated 3D bounding boxes across five object categories: sedan, bus or truck, pedestrian, motorcycle, and bicycle.
Given the object distribution, we focus our evaluation on the sedan and bus/truck classes. For the region of interest (RoI), we use [0, 72] m in the x-axis, [–16, 16] m in the y-axis, and [–2, 7.6] m in the z-axis.
Following the K-Radar benchmark protocol, we evaluate Average Precision (AP) at a 0.3 IoU threshold. Inference speed is reported in frames per second (FPS) to assess the runtime efficiency of our framework.

\subsection{Impact of Proposed Approach}
To demonstrate the effectiveness of the proposed 4DR-MR framework, we conduct a series of experiments focusing on three key aspects:
(1) validating its ability to enhance detection performance under extremely sparse input conditions,
(2) evaluating the effectiveness of incorporating diverse 4DRT representations through multi-teacher KD, and
(3) verifying its overall competitiveness compared to existing 4D Radar-based object detection methods.

\subsubsection{Comparison with RTNH Baselines using Various Pre-processing Techniques}

To support a clearer understanding of our framework's design and the role of diverse data representations, we compare the performance of the networks employed as student and teacher models in our 4DR-MR framework. As shown in Table \ref{tab:RTNH_baseline}, the teacher models are based on RTNH architectures trained on point clouds generated from different 4D Radar pre-processing techniques, including fixed percentile filtering (with thresholds of 90 and 80), interpolation-based percentile filtering, and CA-CFAR. The student model, in contrast, is trained on point clouds generated using only the 99.9th percentile setting.

Despite using the same sparse input as RTNH${99.9}$, 4DR-MR achieves substantial improvements—boosting detection performance by 7.3\% for Sedan and 6.1\% for Bus or Truck in terms of $AP{3D}$, and by 9.5\% for Sedan and 3.42\% for Bus or Truck in terms of $AP_{BEV}$. These results demonstrate the effectiveness of our multi-representation distillation approach even under extremely sparse Radar input.
 Moreover, compared to RTNH$_{90}$, which shows comparable detection performance, the proposed 4DR-MR reduces the per-frame input size by approximately 90 times, highlighting its advantage in terms of both lightweight efficiency and detection accuracy.

\begin{table}[]
\centering
\caption{Student performance comparison on the Sedan class using single vs. multi-teacher KD. Multiple diverse 4DRT-based teachers enhance representation learning. \textbf{Bold} indicates the best result in each metric.}
\label{tab:teacher}
\renewcommand{\arraystretch}{1.2} % 행 간격 조정
\setlength{\tabcolsep}{8pt} % 열 간격 조정
\resizebox{0.8\columnwidth}{!}{%
\begin{tabular}{c|cc|cc}
\hline \hline
\multirow{1}{*}{Approach}       & \multicolumn{2}{c|}{\multirow{1}{*}{Teacher}} & \multirow{1}{*}{$AP_{3D}$}  & \multirow{1}{*}{$AP_{BEV}$} \\ \hline \hline
\multicolumn{1}{c|}{Baseline} & \multicolumn{2}{c|}{N/A}             & 36.84  & 43.10            \\ \hline
\multirow{3}{*}{Single-teacher} & \multicolumn{2}{l|}{(1) RTNH$_{80}$}             & \textbf{44.52}   & 47.79       \\
                              & \multicolumn{2}{l|}{(2) RTNH}        & 44.48      & 47.52                \\
                              & \multicolumn{2}{l|}{(3) RTNH$_{cfar}$}  & 44.04  & 47.22         \\ \hline
\multirow{2}{*}{Multi-teacher}  & \multicolumn{2}{l|}{(1), (2), (3)}            & 44.34            & 47.21    \\
                              & \multicolumn{2}{l|}{(1), (1-1), (2)} & 44.16  & \textbf{52.59}          \\ \hline \hline
\end{tabular}%
}
\end{table}

\subsubsection{Effectiveness of Teacher Diversity}
To validate the benefit of using diverse 4DRT representations, we compare student performance under single-teacher and multi-teacher KD. As shown in Table~\ref{tab:teacher}, the student model trained with multiple teachers—each using a different pre-processing technique—achieves higher AP scores, notably reaching 52.59 for Sedan, demonstrating the effectiveness of integrating complementary features.

However, not all combinations of teachers yield equal gains. For example, the combination of (1), (2), and (3) shows lower performance than (1), (1-1), (2). One possible explanation is that the CA-CFAR-based teacher (3) tends to focus on strong edge reflections and may lack dense object interior information, which could limit its contribution when combined with other teachers.

This suggests that the diversity of representations alone is not sufficient—teacher compatibility and complementarity also play an important role in distillation performance.

\begin{table}[]
\centering
\caption{Comparison with other 4D Radar-based detection networks on the Sedan class. \textbf{Bold} indicates the best result in each metric.}
\label{tab:4DR_OD}
\renewcommand{\arraystretch}{1.2} % 행 간격 조정
\setlength{\tabcolsep}{8pt} % 열 간격 조정
\resizebox{0.7\columnwidth}{!}{%
\begin{tabular}{c|cc}
\hline \hline
\multirow{1}{*}{Network} & $AP_{3D}$         & $AP_{BEV}$       \\ \hline \hline
RTNH\cite{kradar}                     & 36.84          & 43.10           \\ \hline
Radar PillarNet\cite{rcfusion}                & 37.58          & 46.02          \\ \hline
RPFA\cite{rpfa}                     & 37.49          & 40.00           \\ \hline
SMURF\cite{smurf}                & 37.57          & 40.01         \\ \hline
\textbf{4DR-MR}                   & \textbf{44.16} & \textbf{52.59}      \\ \hline \hline
\end{tabular}%
}
\end{table}

\subsubsection{Comparison with Other 4D Radar-based Detection Networks}
To demonstrate the general effectiveness of the proposed 4DR-MR framework, we compare its performance against existing 4D Radar-based 3D object detection networks. As shown in Table \ref{tab:4DR_OD}, 4DR-MR achieves superior performance not only compared to the RTNH baseline, but also when evaluated against other state-of-the-art methods.

While 4DR-MR shows the best overall performance, its results on the Bus or Truck class remain comparable to prior works. This may stem from the RTNH baseline’s relatively lower performance on large vehicles, which affects downstream improvements despite overall gains.

\subsection{Ablation Study}

\begin{table}[]
\caption{Comparison of different fusion strategies for integrating multi-teacher features on the Sedan class. \textbf{Bold} indicates the proposed method.}
\label{tab:aggregation}
\centering
\renewcommand{\arraystretch}{1.2} % 행 간격 조정
\setlength{\tabcolsep}{8pt} % 열 간격 조정
\resizebox{0.7\columnwidth}{!}{%
\begin{tabular}{c|cc}
\hline \hline
\multirow{1}{*}{Fusion Method}   & $AP_{3D}$         & $AP_{BEV}$        \\ \hline \hline
Learnable query-based          & 36.84          & 43.10          \\ \hline
CBAM                           & 37.58          & 46.02          \\ \hline
\textbf{Aggregation module}             & \textbf{44.16} & \textbf{52.59} \\ \hline \hline
\end{tabular}%
}
\end{table}

\subsubsection{Aggregation Module}
We compare several feature fusion techniques to validate the effectiveness of the proposed two-block Aggregation Module. As shown in Table \ref{tab:aggregation}, naive fusion methods fail to distinguish between the roles of teacher-specific alignment and multi-teacher feature selection, resulting in suboptimal performance. Notably, the performance of these alternatives falls below that of single-teacher distillation, highlighting the importance of explicitly separating feature alignment and attention-based fusion. Our proposed two-block Aggregation Module, which explicitly separates representation alignment and attention-based fusion, achieves the highest performance, demonstrating a more balanced and effective integration of diverse 4DRT-derived features.

\subsubsection{Analysis of Qualitative Results}
\begin{figure}[t]
    \centering
    \includegraphics[width=1.0\linewidth]{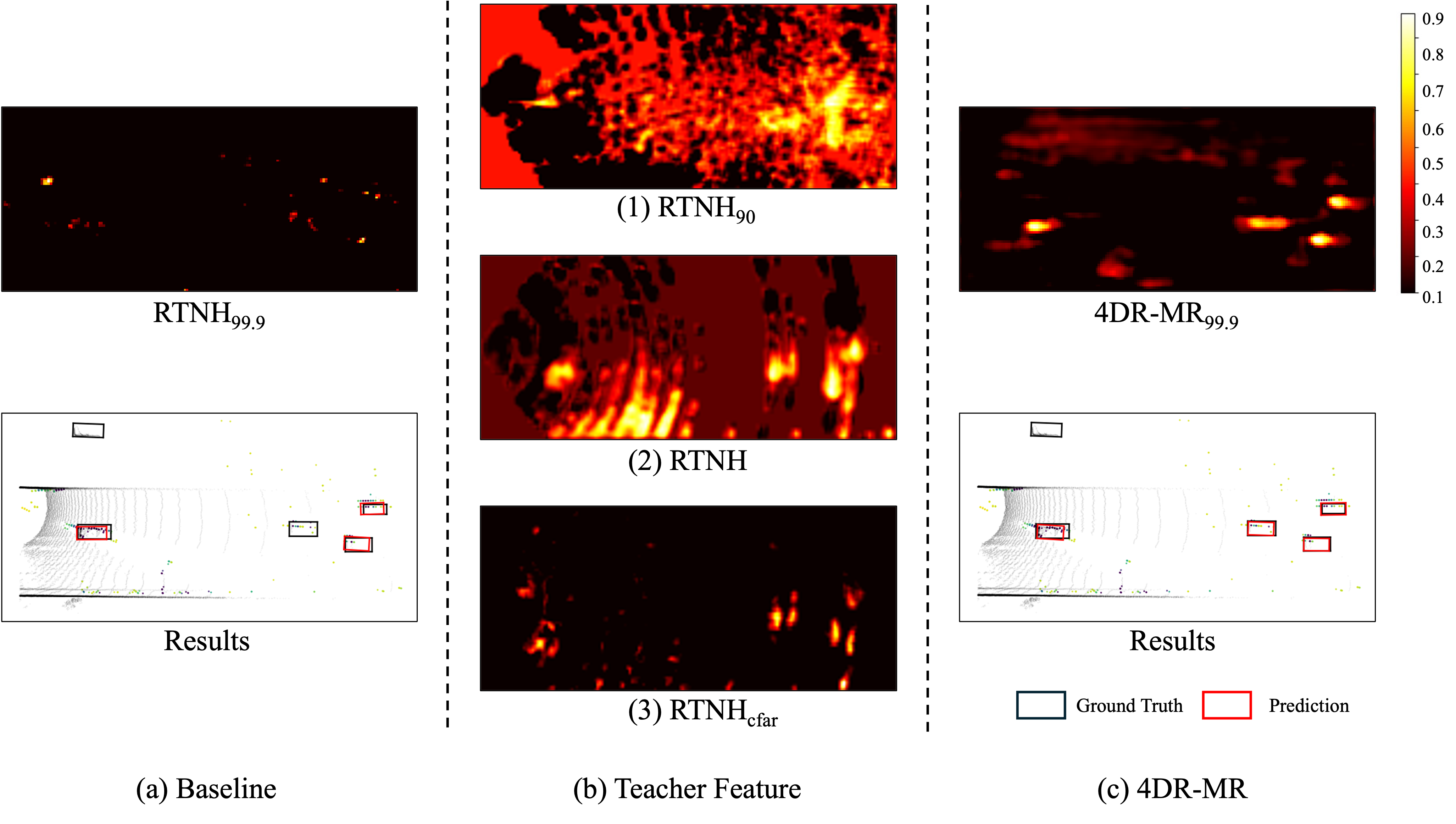}
    \caption{Qualitative comparison of BEV feature maps and detection results. (a) baseline RTNH$_{99.9}$; (b) teacher models trained on (1) RTNH$_{80}$, (2) RTNH with interpolation, and (3) RTNH$_{cfar}$; (c) 4DR-MR student model.}
    \label{fig:qualitative}
\end{figure}

To qualitatively assess the feature enhancement achieved by the proposed 4DR-MR framework, we visualize and compare the BEV feature maps of the baseline RTNH, the teacher models, and the student model. As shown in Fig.~\ref{fig:qualitative}, RTNH$_{99.9}$—trained on extremely sparse Radar inputs—fails to highlight object-relevant regions effectively. In contrast, teacher models trained on richer 4DRT representations produce more salient and diverse features, each capturing complementary aspects of the scene. Guided by this diversity, the 4DR-MR student model—despite using the same sparse input as RTNH$_{99.9}$—generates more informative and semantically consistent feature maps. These results demonstrate that the proposed framework successfully leverages the latent information in 4DRT through multi-teacher KD, enabling robust feature learning under highly sparse input conditions.

\section{Conclusion}
To address the inherent variability and inconsistency of 4D Radar signals—which pose significant challenges for robust perception—we proposed 4DR-MR, a novel 3D object detection framework fundamentally different from prior approaches. By leveraging multiple pre-processing techniques to extract diverse representations from 4DRT, our multi-teacher KD enables a lightweight student model to learn rich features from sparse inputs. Through the proposed aggregation module and efficient distillation process, our method effectively bridges the gap between raw data utilization and resource efficiency. Experimental results on the K-Radar dataset validate the effectiveness of 4DR-MR, demonstrating superior detection performance with high computational efficiency. 

Our findings highlight the necessity of embracing diverse signal representations in Radar perception, and point to future research on efficient methodologies for leveraging 4DRT in a resource-conscious yet accurate manner.

\appendices
\section{Detailed Network Architecture}
\label{appendix:architecture_detail}

\subsection{Aggregation Module Implementation}
Each teacher feature map is passed through a separate alignment block designed to normalize feature semantics prior to fusion.
The alignment block begins with a $1\times1$ convolution that reduces the channel dimension from 768 to 256, followed by batch normalization and ReLU activation.
Next, a ConvNeXt-style block is applied, consisting of three consecutive $1\times1$ convolutions with intermediate LayerNorm and GELU activations. This block expands the channel dimension to 768 and then reduces it back to 256.
A final $1\times1$ convolution and batch normalization reduce the feature to a common embedding dimension of 128, followed by ReLU activation.

The aligned features from each teacher are concatenated along the channel dimension and passed through a Convolutional Block Attention Module (CBAM) to emphasize informative regions.
CBAM comprises two components:
\begin{itemize}
    \item \textbf{ChannelGate:} Applies average and max pooling across spatial dimensions, followed by a two-layer MLP with reduction ratio $r=8$.
    \item \textbf{SpatialGate:} Applies average and max pooling across channels, concatenates the results, and applies a $7\times7$ convolution for spatial attention.
\end{itemize}
Finally, a $1\times1$ convolution is applied to produce the aggregated teacher feature.

\subsection{Densify Module Implementation}
To densify sparse student features and reduce mismatch with dense teacher features, we apply a two-stage encoder–decoder architecture extended from \cite{sparse2dense}.
Each stage consists of:
\begin{itemize}
    \item A ConvNeXt-style encoder: sequential residual blocks with downsampling to capture global context.
    \item A decoder: transposed convolutions to restore spatial resolution.
\end{itemize}
The student backbone feature is first passed through the first encoder–decoder block to produce an intermediate dense representation.
This is then passed through a second encoder–decoder block to further refine feature quality.
The output of the second decoder is fused with the original student feature via a residual connection.
Finally, the fused representation is processed by a sequence of convolutional layers to yield the final student feature $F_S$.

\bibliographystyle{IEEEtran}
\bibliography{ref}

\end{document}